\documentclass[letterpaper, 10 pt, conference]{ieeeconf}     
\usepackage{cite}
\usepackage{todonotes}
\usepackage{amsfonts}
\usepackage{amsmath}
\usepackage{graphicx}
\usepackage{subcaption}
\usepackage{notoccite}
\usepackage{hyperref}
\usepackage{url}
\usepackage{epsfig} 
\usepackage{mathptmx} 
\usepackage{times} 
\usepackage{amssymb}  
\usepackage{multirow}
\usepackage[normalem]{ulem}
\useunder{\uline}{\ul}{}
\usepackage{mathtools}
\usepackage{cite}

\IEEEoverridecommandlockouts                              
\hypersetup{
    colorlinks = true,
    linkcolor = {blue},
    citecolor = {blue}
}

\graphicspath{{./Images/}}

\title{\LARGE \bf
``Touching to See" and ``Seeing to Feel": Robotic Cross-modal Sensory Data Generation for Visual-Tactile Perception
}

\author{Jet-Tsyn Lee, 
Danushka Bollegala 
and Shan Luo
\\
\thanks{Department of Computer Science, University of Liverpool, Liverpool L69 3BX, U.K.
Emails: \{j.lee35, danushka, shan.luo\}@liverpool.ac.uk.}%
}

\begin{document}

\maketitle
\thispagestyle{empty}
\pagestyle{empty}

\begin{abstract}
The integration of visual-tactile stimulus is common while humans performing daily tasks. In contrast, using unimodal visual or tactile perception limits the perceivable dimensionality of a subject. However, it remains a challenge to integrate the visual and tactile perception to facilitate robotic tasks. In this paper, we propose a novel framework for the cross-modal sensory data generation for visual and tactile perception. Taking texture perception as an example, we apply conditional generative adversarial networks to generate pseudo visual images or tactile outputs from data of the other modality. Extensive experiments on the ViTac dataset of cloth textures show that the proposed method can produce realistic outputs from other sensory inputs. We adopt the structural similarity index to evaluate similarity of the generated output and real data and results show that realistic data have been generated. Classification evaluation has also been performed to show that the inclusion of generated data can improve the perception performance. The proposed framework has potential to expand datasets for classification tasks, generate sensory outputs that are not easy to access, and also advance integrated visual-tactile perception. 

\end{abstract}

\section{Introduction}

Vision and touch are two key information sources for us humans to perceive the physical world. Vision observes the environment by identifying features such as shape, and/or colours. Whilst touch uses the somatosensory system to measure a variety of pressure responses, detecting intricate surfaces and texture of an object. The use of vision or touch individually usually limits our perception ability by restricting observations to a single view \cite{spence_crossmodal_2011}. A combination of senses (vision and touch in this case), known as \textit{multisensory integration}, increases the dimensionality of the observation by capturing an object from multiple viewpoints, reinforcing our interpretation of the environment \cite{calvert_crossmodal_2001}. In our daily experience, we can cognitively create a visualisation of an object based on a tactile response, or a tactile response from viewing a surface's texture \cite{banati_functional_2000}. This perceptual phenomenon, \textit{synaesthesia}, in which the stimulation of one sense causes an involuntary reaction in one or more of the other senses, can be employed to make up an inaccessible sense. For instance, when one grasps an object our vision will be obstructed by the hand but a touch response will be generated to ``see" the corresponding features. This can be common due to the nature of the environment where a perception source can become unavailable at an unexpected time. This is experienced as ``touching to see" or ``seeing to feel" where we interpret features from an alternate perspective based on the information received from a different view \cite{newell_visual_2005}.

\begin{figure}
\centering
\begin{subfigure}{0.5\textwidth}
      \centering
      \includegraphics[width=\textwidth]{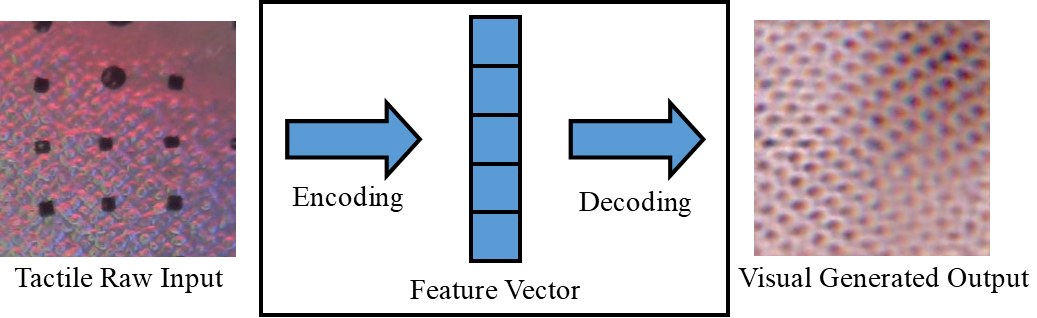}
      \caption{Generation of a tactile-to-visual image}
      \label{fig:11_real}
\end{subfigure}
\begin{subfigure}{0.5\textwidth}
      \centering
      \includegraphics[width=\textwidth]{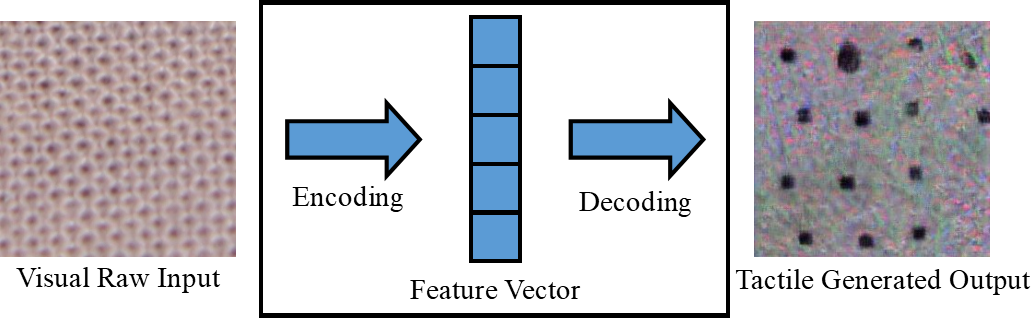}
      \caption{Generation of a visual-to-tactile reading}
      \label{fig:11_cont}
\end{subfigure}
\caption{Cross-modal sensory data generation for visual-tactile perception.}
\label{fig:conv}
\end{figure}

Vision and touch sensing have also been widely applied in robotics. Visual perception can be found in tasks such as object detection, texture recognition and localisation \cite{csurka_domain_2017,peel2018localisation}. And the use of touch sensing for perception of object properties has also been on the rise in recent years as different types of tactile sensors are developed \cite{dahiya_tactile_2010}. Most existing works utilise vision or tactile information individually to perform such tasks, which may limit the sensing capabilities for operating in unstructured environments \cite{luo_2017_robotic,luo2018iclap}. Vision and touch modalities provide complementary information and combining the two modalities could have a synergistic effect, where features in the environment can be perceived in situations where it would be impossible using the information from each sensory channel separately. However, it remains a challenge for models to understand the relationship between different perspectives of objects from vision and touch as there is an infinite number of different views of objects captured from both senses. Visual and tactile views consist of separate sensory domains presented in different feature sets resulting in a performance drop when the model is presented in new domains, known as \textit{dataset shift}, where transferring between views cause issues due to the differences in the distribution of inputs in the training and testing stages \cite{shimodaira_2000_improving}.

To advance the integrated visual-tactile perception, in this paper, we propose a novel framework for cross-modal sensory data generation, by predicting sensory outputs of one sense from data of the other. We take the texture perception as an example: visual input images of a cloth texture are used to generate a pseudo tactile reading of the same piece of cloth; conversely, tactile readings of a cloth are employed to predict a visual image of the same cloth, both concepts shown in Figure \ref{fig:conv}. The textures of fabrics, i.e., the yarn distribution patterns, appear similarly in a visual image and a pressure distribution (tactile) reading. Nevertheless, this work can be extended to the cross-modal visual-tactile data generation for perception of other object properties by taking considerations of the differences in the two domains. As a pioneering work in the robotic cross-sensory data generation, the proposed framework is of utmost importance in the following perspectives: 1). By incorporating generated pseudo sensory data, the datasets for each sensory channel can be expanded for tasks like classification; 2). The sensory data that are not easy to access can be generated from the data of the other modality that is more available; 3). The proposed framework will also support the integration of visual and tactile information for better coordination.


The remainder of this paper is structured as follows: Section \ref{chap:lit} introduces the related works; Section \ref{chap:met} details the cross modal data generation model structures; Section \ref{chap:vit} introduces the GelSight sensor and ViTac visual-tactile dataset of cloth textures; Section \ref{chap:exp} shows the experiment results and analysis; 
Finally, Section \ref{chap:conc} summarizes the conclusions of the paper and describes potential future directions.

\section{Related Works}\label{chap:lit}
The cross-modal visual-tactile data generation lies in the tasks of domain adaptation and cross-modal analysis. 
To this end, in this section, we review the related works in domain adaptation and cross-modal analysis, 
and also works on visual-tactile perception.

\subsection{Domain Adaptation and Cross-modal Analysis}
Domain Adaptation or Transfer Learning algorithms adapt inputs from the source dataset to relate to a target data distribution \cite{pan_survey_2010}, mitigating degradation and dealing with the differences in the data distributions across different domains. Learning to adapt different domains increases the data range as information can be borrowed from similar domains to learn a classifier for unseen or unlabelled data in a target domain. It has been widely applied in the computer vision tasks \cite{ganin_domain-adversarial_2015}. For instance, in \cite{saenko_adapting_2010}, cross-domain transformations are learned for comparing product images against real-world images captured by users. Different from transformations between different domains of one modality, cross-modal analysis aims to learn a common subspace from representations of different modalities and the learned subspace can be projected back to the spaces of different modalities. Many works exist in cross modal retrieval that aim to enable flexible retrieval experience across different modalities, e.g., texts vs. images \cite{wang_2017_adversarial}. In these works, joint representations and correlations of multiple modalities are learned so that samples in the database across different modalities can be retrieved. The cross-modal generation makes a step further, which not only learns the subspace but also generates new data across different modalities, mapping from one modality space to the other. It generates novel images that are unseen, or data in other modalities that are not easy to access. In \cite{isola_image--image_2017}, conditional Generative Adversarial Networks (cGANs) \cite{mirza_conditional_2014} are applied to convert a sketch to generate an image representation. Reed \textit{et al.} \cite{reed_generative_2016} generate plausible images conditioned on detailed text captions. In \cite{chen_2017_deep}, cross-modal audio-visual generation of musical performance is achieved, i.e., generating audio from musical performance images and also vice versa. To the authors' best knowledge, our work is the first attempt to the robotic cross-modal visual-tactile data generation, which can also be extended to cross-modal data generation for other modalities.

\subsection{Visual and Tactile Sensing}
Vision and touch sensing are two main important modalities in perception. Both have been widely applied in robot tasks, usually with only one modality used \cite{luo_2017_robotic, luo2016iterative,bimbo2016hand}. It is still challenging to combine vision and touch modalities to facilitate robot operations due to their different sensing principles and data structures. In \cite{kroemer_2011_learning}, vision and tactile samples are paired to classify materials using dimensionality reduction techniques. In \cite{luo2015localizing}, tactile contacts are localized in a visual map by matching the tactile features with visual features. Vision and touch data are combined to reconstruct a point cloud representation and there is no learning of the key features of the two modalities in \cite{izatt_2017_tracking}. Deep neural networks have also been used to extract adjectives/features from both vision and tactile data \cite{gao_2015_deep, luo_vitac:_2018}. In a more recent work \cite{falco_cross-modal_2017}, a cross-model framework is proposed for visuo-tactile object recognition. Differently from the prior works on learning a subspace of vision and touch, we make a step further to generate new tactile-visual data.



\section{Methodology}\label{chap:met}
In this section, our cross-modal visual-tactile data generation models are described, i.e., the cross-modal adaptation model structures, learning algorithms, and the evaluation metrics.

\subsection{Visual-Tactile Data Generation Models}
\begin{figure}
\begin{subfigure}{0.5\textwidth}
  \centering
  \includegraphics[width=\textwidth]{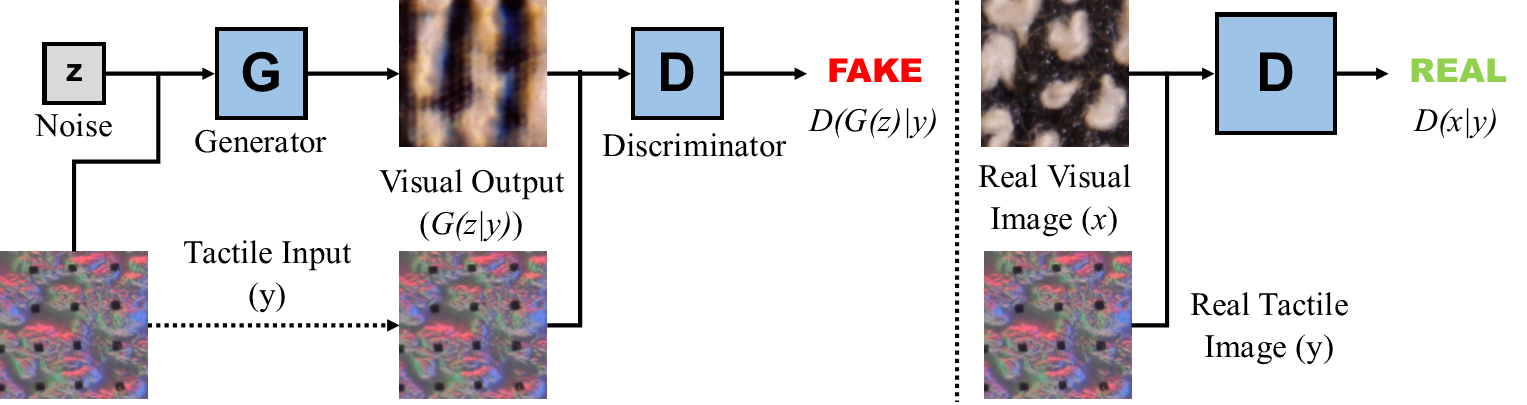}
  \caption{Tactile-to-Visual Network}
  \label{fig:T2V}
\end{subfigure}
\begin{subfigure}{0.5\textwidth}
  \centering
  \includegraphics[width=\textwidth]{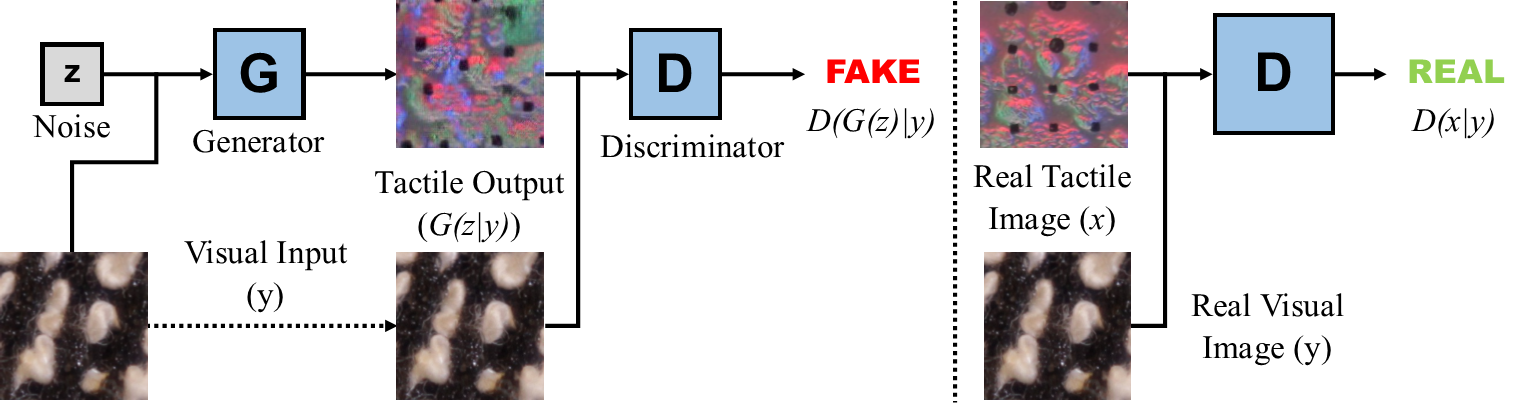}
  \caption{Visual-to-Tactile Network}
  \label{fig:V2T}
\end{subfigure}
\caption[Conditional Generative Adversarial Network]
{Diagram of the cross-modal visual-tactile generation model which consists of (a) tactile-to-visual network and (b) visual-to-tactile network showing the generation process and the classification from the discriminator.}
\label{fig:cgan}
\end{figure}

The goal of this paper is to learn the relationship between visual and tactile perception by building a model that understands the cross-modal interaction between the two views.
For our robotic application, we explore the use of cGANs \cite{mirza_conditional_2014}, a variant of Generative Adversarial Networks \cite{goodfellow_generative_nodate}, to learn the relationship between the two domains to generate artificial data in the opposing view. We use an auxiliary information $Y$ to represent the input domain in our network and $X$ as the target domain. 
The layout of the cross-modal visual-tactile data generation is illustrated in Figure \ref{fig:cgan}. 
The model aims to learn the mapping between the domains $X$ and $Y$ given training samples $\{x_i\}^N_{i=1}$ where $x_i\in X$ and $\{y_j\}^M_{j=1}$ where $y_j\in Y$.
This conditions the output giving a higher quality image and more control in the network by manipulating $Y$ in both $G$ and $D$ as an additional layer, to convert the input image of our initial domain to our target output.
The objective function can be expressed as:
\begin{multline}
\min_{G}\max_{D}V(D,G)= 
\mathbb{E}_{x \sim p_{data}(x)} [\log D(x|y)] + \\
\mathbb{E}_{z \sim p_z (z)}[\log (1-D(G(z|y)))]
\end{multline}
where $G(z|y)$ tries to generate an image similar to the $X$ domain with $z$ as an added noise vector to the input.
The concept of the model is shown in Figure \ref{fig:cgan}.
In the case of the tactile-to-visual generation, as shown in Figure \ref{fig:T2V}, the tactile image is applied along with a noise vector to convert the image to the visual domain.
We then classify the generated image through the discriminator which determines the image as fake to update the internal network.
Real images are also classified along with the generated image for the model to improve the classification ability of the network. The visual-to-tactile generation is similar to the tactile-to-visual generation, as shown in Figure \ref{fig:V2T}.

\subsection{Network Structure}
Our generator and discriminator are adapted from \cite{isola_image--image_2017}.
The generator architecture uses an auto-encoder network to encode the input image to a feature vector and allows the decoder to reverse the process and recreate the image to the opposing view.
We extend the architecture to a U-Net applying up-sampled layers between the encoding, $E_i$, and decoding, $D_{(n-1)}$, bypassing unused layers to improve the performance and allow the network to work with fewer training images \cite{ronneberger_u-net:_2015}. 
Both the generator and discriminator apply convolution-BatchNorm-ReLu from \cite{ioffe_batch_2015} using 7 layers repeating the actions for 256$\times$256 images.
Noise $z$ is provided in terms of a random uniform positioning of the images that randomizes the feature positions in the same latent space.

\subsection{Optimization}
The cross-entropy error is computed using the probabilities returned by a logistic sigmoid function, a standard approach from \cite{goodfellow_generative_nodate}. 
This predicts a probability value between $[0,1]$, where $1$ indicates the real input, and zero otherwise. 
This results in $G$ aiming to minimise the objective against the adversary $D$ that tries to maximize the function.
The network is optimized using an RMSProp as an alternative to a stochastic gradient descent, letting us use a larger learning rate $\alpha$ without risking premature convergence \cite{ronneberger_u-net:_2015}.

\subsection{Evaluation Metric}
Due to the lack of an objective function in GANs structures, it is difficult to evaluate the quality of the generated images and compare the performance of the model. 
The most effective way is to use human annotators to visually distinguish between the real and fake images. 
Services such as Amazon Mechanical Turk allow users to evaluate the outputs of the network, however, feedback is highly reliant on the annotator with a varying degree of results. Furthermore, in our case, the generated tactile outputs are not appropriate to be evaluated by visual inspection in such services. Instead, we employ two evaluation methods to measure the quality of the generated outputs and the real dataset as a comparison: one to use structural similarity index to compare the pseudo and real images, and one to test whether the inclusion of generated data will increase the recognition performance.



\begin{figure*}[!h]
\centering
\begin{subfigure}{0\textwidth}
\refstepcounter{subfigure}\label{fig:c22}
\end{subfigure}%
\begin{subfigure}{0\textwidth}
\refstepcounter{subfigure}\label{fig:c23}
\end{subfigure}
\begin{subfigure}{0\textwidth}
\refstepcounter{subfigure}\label{fig:c100}
\end{subfigure}
\includegraphics[width=\textwidth]{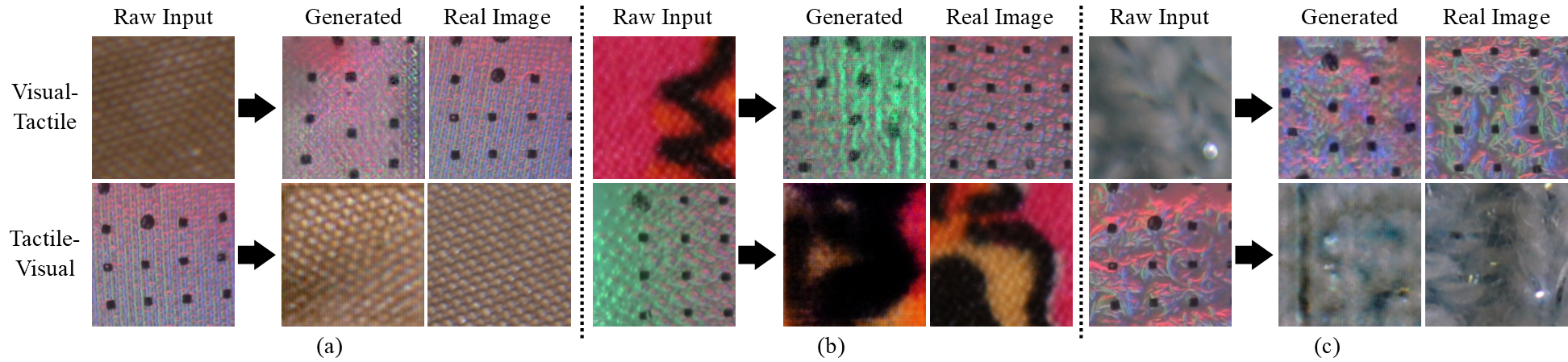}
\caption{Images showing several generated outputs from different cloth materials converting an input image between a visual and a tactile view (a) Ideal training images generating a consistent pattern (b) Cloth set containing a dyed design showing difficulty in replicating a tactile-visual image (c) Wool fabric with a low quality image generated.}
\label{fig:cloth_test}
\end{figure*}

We adopt the Quaternion-SSIM \cite{kolaman_quaternion_2012} and have the Colour Structural Similarity (Colour-SSIM) index as follows, to measure the image quality under the 3 colour channels, $RGB$, to capture the full colour spectrum and calculate the similarity for each channel:
\begin{multline}
\operatorname{Colour-SSIM}(x,y)= \\
\frac{(2\mu_x^{(i)} \mu_y^{(i)} +C_1)(2\sigma_{xy}^{(i)} +C_2)}
{((\mu_x^{(i)})^2+(\mu_y^{(i)})^2+C_1)((\sigma_x^{(i)})^2+(\sigma_y^{(i)})^2+C_2)}
\end{multline}
where $\mu_x$ and $\mu_y$ are the averages for variables $x$ and $y$, $\sigma_x$, $\sigma_y$ are the variance for $x, y$, respectively, and $\sigma_{xy}$ is the covariance of $x$ and $y$; $C_1=(k_1L)^2$, $C_2 = (k_2L)^2$ are added constants for numerical stability with $L$ as the pixel-value dynamic range, and $k_1=0.01$, $k_2=0.03$ as a scalar default, and $i$ represents the different colour channels.

\section{ViTac Dataset}\label{chap:vit}
The ViTac Cloth dataset \cite{luo_vitac:_2018} consists of a collection of 100 different pieces of fabrics with a variety of different materials and textile patterns such as knitted cotton, denim, or dyed polyester. 
The dataset contains two types of data per cloth type: 
\begin{itemize}
\item Visual macro images captured from a Canon T2i SLR camera to represent the visual perception of the material with 1,000 digital camera images in the ViTac dataset;
\item Tactile readings captured using a GelSight sensor with 96,536 tactile images collected.
\end{itemize}

The GelSight sensor used to collect the tactile data measures the geometry of a surface by physically contacting an object. 
The sensor face is covered with a clear elastomer gel coated with a reflective membrane. The object in contact deforms the gel and takes shape of the surface. A camera underneath the gel captures frames to record the reflective surface showing the deformation of the surface. The membrane is illuminated by LEDs that project RGB lights in different directions to measure the intensity of the force applied. 
The gel surface is marked with a grid pattern to indicate the direction of the force applied by capturing the changes of the position per frame. More details of the dataset and the sensor can be found in our previous work \cite{luo_vitac:_2018}.
This dataset represents the visual and tactile perception of the application for the model to understand the relationship of the two distributions. By perceiving a similar view with both devices, we present the perception to the model to identify the related features and reconstruct the fabric texture in the opposing view.

\section{Experiments}\label{chap:exp}
We conduct several experiments to measure the properties of the cross-modal data generation models for generating the desired output.
We trained the visual-to-tactile and tactile-to-visual networks to generate tactile and visual images against a number of materials with different fabric properties, and altering the networks internal parameters and input. Outputs are evaluated against a Colour-SSIM metric comparing the generated artificial image against the real dataset. 
Our code is implemented in Python using the Tensorflow framework\footnote{https://www.tensorflow.org/} and all experiments are computed using the University of Liverpool GPU server running on 3$\times$Nvidia GeForce GTX 1080 Titan GPUs.

\subsection{Network training for generating images}

\subsubsection{Cloth Properties}
Figure \ref{fig:cloth_test} shows the generated outputs against different cloth types to understand the capabilities of adapting different material properties to replicate the real dataset. 
We conduct individual tests to convert visual-to-tactile and tactile-to-visual images to understand both viewpoints. 
The selected cloth types for the experiments were chosen to highlight the different properties of the materials, e.g., weaving patterns, colour, and material type, to test the model's generation process in adapting different materials. 

\begin{figure*}
\centering
\begin{subfigure}{0\textwidth}
\refstepcounter{subfigure}\label{fig:mreal}
\end{subfigure}%
\begin{subfigure}{0\textwidth}
\refstepcounter{subfigure}\label{fig:mdef}
\end{subfigure}
\begin{subfigure}{0\textwidth}
\refstepcounter{subfigure}\label{fig:mbatch}
\end{subfigure}\begin{subfigure}{0\textwidth}
\refstepcounter{subfigure}\label{fig:ml1}
\end{subfigure}%
\begin{subfigure}{0\textwidth}
\refstepcounter{subfigure}\label{fig:miter}
\end{subfigure}
\begin{subfigure}{0\textwidth}
\refstepcounter{subfigure}\label{fig:mdata}
\end{subfigure}
\begin{subfigure}{0\textwidth}
\refstepcounter{subfigure}\label{fig:mgt}
\end{subfigure}
\begin{subfigure}{0\textwidth}
\refstepcounter{subfigure}\label{fig:mres}
\end{subfigure}
\begin{subfigure}{0\textwidth}
\refstepcounter{subfigure}\label{fig:ROI}
\end{subfigure}
\begin{subfigure}{0\textwidth}
\refstepcounter{subfigure}\label{fig:mix}
\end{subfigure}
\includegraphics[width=\textwidth]{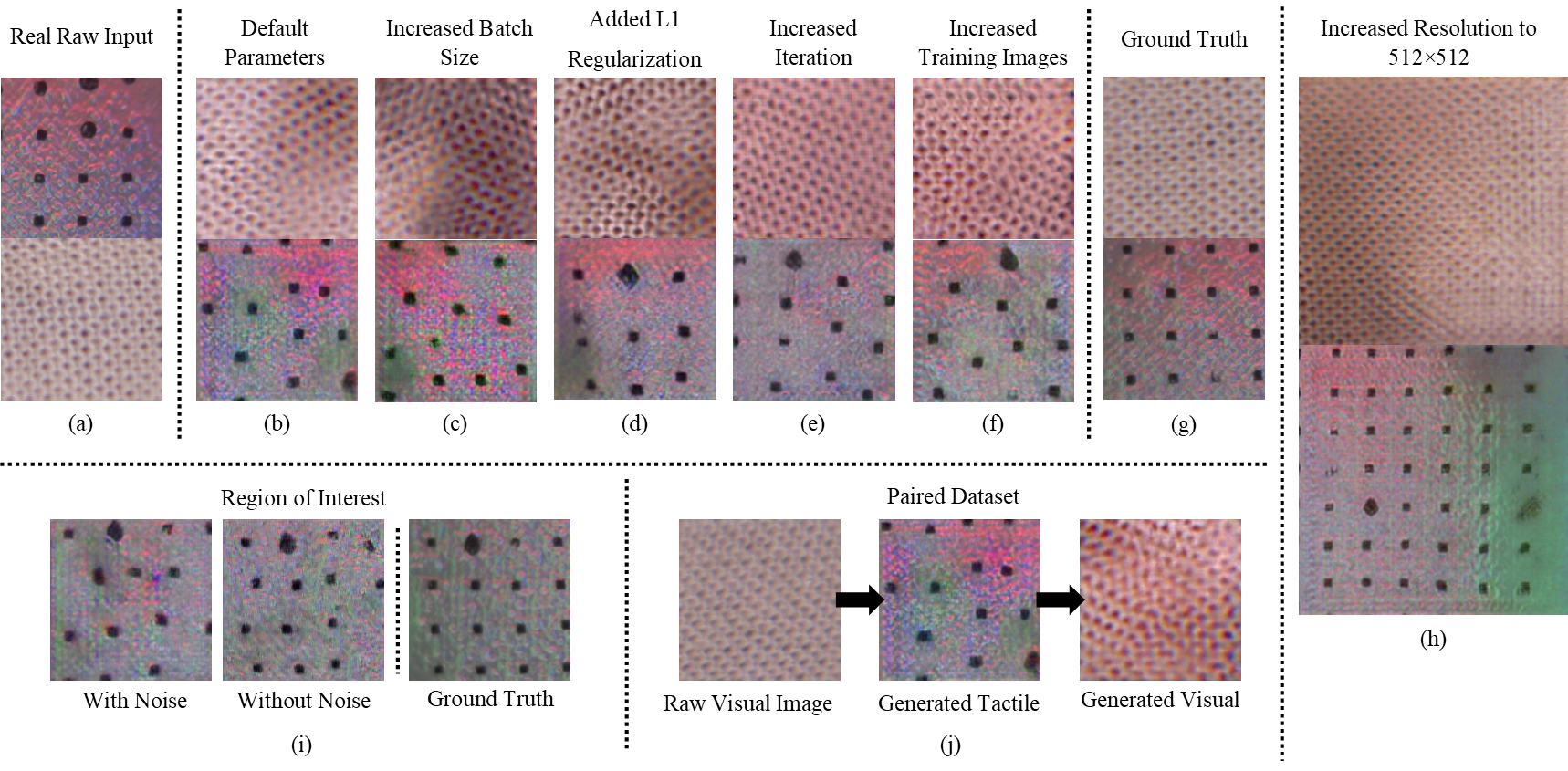}
\caption{Generated outputs from changing the network parameters. Top: Tactile-to-Visual; Bottom: Visual-to-Tactile. (a) Input Images (b) Outputs trained using default parameters (c) Increased batch size applied for each iteration  (d) Added L1 Regularization to generator loss (e) Increased iteration amount before termination (f) Increased amount of training images (g) Ground Truth (h) Increased image resolution from 256$\times$256 to 512$\times$512 (i) Trained with the dataset consisting of a ROI (j) Trained to reverse the domain generation.}
\label{fig:modelgen}
\end{figure*}

\subsubsection{Network Parameters}
We analyse the generation process by adjusting internal parameters and inputs in identifying the behaviour and actions to improve the generated image quality. 
The same cloth type is used for the experiments to ensure that no factors from the training images impact the results and to measure the changes in the output. 
Tests conducted to convert images from visual-to-tactile and tactile-to-visual domains. 
We train the model from default parameters to measure the changes from further tests altering the parameters individually from the default setting including increasing the batch size, iterations, and adding an L1 regularization to the generator loss \cite{pathak_context_2016}.
This also includes alterations to the image input such as increasing the number of training images used for each domain and increasing the resolution size of the input image 512$\times$512. 
Figure \ref{fig:modelgen} shows the resulting generated output from the test.

\subsubsection{Region of Interest}
Figure \ref{fig:ROI} shows the generated tactile image of an exclusive dataset by selecting a Region of Interest (ROI) from the GelSight tactile images with a similar image structure.
This aims to reduce views from the tactile domain as we select an identical region with similarities mitigating inconsistencies in the training images.
We trained the network with and without noise to see the effects of the output, and tested only with a visual-to-tactile generation, as the visual data is already similar.

\subsubsection{Paired Dataset}
A final experiment trains the network on a 50/50 set of real and generated images to recreate the outputs into the original domain. 
This is to capture the assumption that if we translate from one domain to the other and back again, we should arrive at where we started.
Outputs of the test are illustrated in Figure \ref{fig:mix}.

\subsection{Evaluating Visual-to-Tactile Generation}
Generation between a visual-to-tactile view ignores any visual features from the input dataset such as the colours and design pattern as seen in Figure \ref{fig:c23}.
This allows us to ignore the colour scheme of the material when converting to a tactile view, however, becomes difficult in classification tasks when trying to determine the difference between the same material with different colours.
Generated images shown in Figure \ref{fig:c22} and Figure \ref{fig:c23} are able to replicate the textile pattern as both visual and tactile datasets capture the features from a consistent pattern to relate the features in both views.
As shown in Figure \ref{fig:ROI}, by minimizing the multiple views and noise, the quality of the generation pattern can be improved as trained with a more consistent dataset.
Low-quality results such as Figure \ref{fig:c100} using a wool material was not able to capture the weaving pattern due to the different views  of the training images, but instead aims to replicate the fibres from the material.
By capturing a larger images range as seen in Figure \ref{fig:mres}, higher resolution will allow the process to capture more features in both perspectives to match the patterns which could improve the generation of difficult materials.

Training time averages $\approx{4}$ hours for most experiments, with the exception of several network parameter tests, such as an increased iteration amount, or image resolution, proportional to the test as expected.
Image generation with the material seen in Figure \ref{fig:c23} requires an $\approx{120\%}$ increase in resources due to the variation of the visual perception. 
This is due to difficulties in relating both views as the colour variation of the fabric proves difficult in identifying the features from the visual perception to relate in a tactile view.

From Table \ref{table:eval_vt} we can see the visual-to-tactile outputs evaluated against the Colour-SSIM metrics.
Resulting outputs gave a similarity of $\approx$0.9 for most outputs as the generated result was able to replicate the key properties of the GelSight sensor data. Although the visual-tactile network was able to generate good outputs, further improvements are possible with further fine-tuning of the parameters to improve the generated results, and improvements made to the training dataset, as seen in Figure \ref{fig:modelgen}.
\begin{table}[!h]
\centering
\begin{tabular}{|c|c|}
\hline
\textbf{Visual-to-Tactile}  & \textbf{Colour-SSIM} \\ \hline
\textit{Cloth Types}        & 0.89717              \\ \hline
\textit{Network Parameters} & 0.90896              \\ \hline
\textit{ROI, with noise}    & 0.92059              \\ \hline
\textit{ROI, no noise}      & 0.91188              \\ \hline
\textit{Paired Dataset}     & 0.90471              \\ \hline
\end{tabular}
\caption[]
{Evaluation results applying the Colour-SSIM metric to measure the generated visual-to-tactile images against the real tactile dataset. Metric was applied to all experiments and the resulting average was taken for the set of experiments.}
\label{table:eval_vt}
\end{table}

\subsection{Evaluating Tactile-to-Visual Generation}
The tactile-to-visual image as seen in Figure \ref{fig:c22} was able to capture the key patterns of the material with inconsistencies due to the different views and rotations of the training images. As shown in \ref{fig:miter}, by increasing the iterations, the inconsistent pattern can be resolved by training a longer period to identify the features from the training images.

Unlike from the visual-to-tactile recognition, the model needs to account for the visual perspective capturing the colour of the material.
Materials such as Figure \ref{fig:c23} had difficulty in recreating the fabric design as the location of the pattern cannot be captured in the tactile view.
Figure \ref{fig:c100} lacked details on the training images due to similar colour variation making it difficult to capture the fabric pattern for the model to relate features in both views. The result shows success in improving the sharpness of the visual output by altering several internal parameters as seen in Figures \ref{fig:mbatch}, \ref{fig:ml1}, and \ref{fig:mdata}.
Increasing the resolution resulted in a better consistent image for both conversions by capturing the full image with a minimal noise applied as the size of the image makes the noise redundant. 

Similar to the Tactile-Visual generation, training time takes $\approx{4}$ hours.
Again, we require higher resources for coloured fabrics seen in Figure~\ref{fig:c23} due to the differences in the visual view, showing challenges in replicating the perception due to the number of variations.  

The evaluation results applying the tactile-to visual outputs are listd in Table \ref{table:eval_tv}. Applying the Colour-SSIM metric against the different cloth types resulted in a varying similarity depending on the fabric type.
Fabrics with visually identical patterns generated such as Figure \ref{fig:c22} resulted in better similarity percentage.
Compared to other generated images such as \ref{fig:c23} resulted in a lower score averaging $\approx$0.65, as the generation was not successful in replicating the fabric pattern resulting in varying outputs which cannot be compared against the real dataset.
\begin{table}[!h]
\centering
\begin{tabular}{|c|c|}
\hline
\textbf{Tactile-to-Visual}  & \textbf{Colour-SSIM} \\ \hline
\textit{Cloth Types}        & 0.77279              \\ \hline
\textit{Network Parameters} & 0.89595              \\ \hline
\textit{Paired Dataset}     & 0.91338              \\ \hline
\end{tabular}
\caption[]
{Evaluation results applying the Colour-SSIM metric to measure the generated tactile-to-visual images against the real visual dataset. Metric was applied to all experiments and the resulting average was taken for the set of experiments.}
\label{table:eval_tv}
\end{table}

\subsection{Classification Evaluation}
We evaluate the generated images by applying the new data in a classification task, using the AlexNet CNN \cite{krizhevsky_imagenet_2017} for our classification and retraining the final layers to be suited for our dataset.
The network is trained on 1,100 images categorised into 11 different classes.
We calculate the accuracies under two instances by only applying the real visual and tactile images, and testing the application including both the real and generated images.
Table \ref{table:class} shows the classification accuracies of both visual and tactile datasets. 

Early in the training, a distinct increase can be seen in the accuracy by applying the generated dataset but eventually converges after a number of iterations.
This shows the benefit of substituting the generated images if the real dataset is limited, however, can be mitigated by training with a longer number of iterations. 
We see marginal improvements to the visual image accuracy by applying the generated images to the task.
The tactile classification resulted in a higher variation due to the similarities in several fabric types using the GelSight image causing some difficulty in correctly classifying several cloth types.

\begin{table}[!h]
\begin{tabular}{c|c|c|c|c|}
\cline{2-5}
\textbf{}                                & \multicolumn{2}{c|}{\textbf{Visual Images}} & \multicolumn{2}{c|}{\textbf{Tactile Images}} \\ \hline
\multicolumn{1}{|c|}{\textbf{Iteration}} & \textbf{Real Images}  & \textbf{Real/Gen.}  & \textbf{Real Images}   & \textbf{Real/Gen.}  \\ \hline
\multicolumn{1}{|c|}{\textit{1}}         & 0.9623                & 0.9710              & 0.7776                 & 0.8506              \\ \hline
\multicolumn{1}{|c|}{\textit{2}}         & 0.9789                & 0.9770              & 0.8249                 & 0.8465              \\ \hline
\multicolumn{1}{|c|}{\textit{3}}         & 0.9802                & 0.9853              & 0.8902                 & 0.8699              \\ \hline
\multicolumn{1}{|c|}{\textit{4}}         & 0.9830                & 0.9839              & 0.8952                 & 0.8736              \\ \hline
\multicolumn{1}{|c|}{\textit{5}}         & 0.9858                & 0.9835              & 0.9118                 & 0.8911              \\ \hline
\multicolumn{1}{|c|}{\textit{6}}         & 0.9821                & 0.9867              & 0.9007                 & 0.8989              \\ \hline
\multicolumn{1}{|c|}{\textit{7}}         & 0.9871                & 0.9867              & 0.9127                 & 0.9067              \\ \hline
\multicolumn{1}{|c|}{\textit{8}}         & 0.9858                & 0.9867              & 0.9141                 & 0.9044              \\ \hline
\multicolumn{1}{|c|}{\textit{9}}         & 0.9876                & 0.9894              & 0.9099                 & 0.8989              \\ \hline
\multicolumn{1}{|c|}{\textit{10}}        & 0.9881                & 0.9894              & 0.9131                 & 0.9058              \\ \hline
\end{tabular}
\caption{Classification accuracy testing with real images only, and both real and generated images for the visual and tactile viewpoints.}
\label{table:class}
\end{table}

\section{Discussion and Conclusion}\label{chap:conc}

In this work, we explore the cross-modal relationship between visual and tactile perception with the use in a robotic application. We employ the use of conditional GANs to learn to adapt visual and tactile data to the opposing domain and conducted several experiments using the ViTac dataset to generate outputs in the opposing domain. This tests numerous cloth properties the model may encounter to understand the ability to adapt the different perceptual views. For the visual-to-tactile generation, significant results were seen in generating key patterns to replicate the tactile view of the GelSight sensor. Fabrics with a consistent pattern were able to be replicated by matching the feature that was able to be captured from the visual and tactile perspectives. The training images contain multiple views having difficulty in recreating the pattern. 
For the tactile-to-visual generation, significant results were also seen in the generated images in recreating the key visual patterns
The network shows more success in generating a tactile-to-visual image as the visual perspective was able to capture the details of the fabric. 
Colour-SSIM index was used to evaluate the generated visual and tactile images, averaged a $\approx$0.9 similarity against the real tactile image for most cases as the images are visually similar to the GelSight tactile images. Evaluation by applying the Colour-SSIM metric against the tactile-to-visual images ranges from $\approx$0.50 to 0.90 similarity depending on the cloth type as several materials were not able to capture the key patterns that can match the real dataset.

Several directions for future work include:
Further tests to the network parameters to further improve the quality of the image generation process, and train the network with transfer learning techniques on a number of different datasets and objects to further expand the capabilities of the network. Additional study may include integrating the Visual-Tactile relationship for a number of different tasks, such as classification in both a visual and tactile setting for a real world setting.

\section*{ACKNOWLEDGMENT}

This work was supported by the EPSRC project ``Robotics and Artificial Intelligence for Nuclear (RAIN)" (EP/R026084/1).


{\small
\bibliography{Libary}{}
\bibliographystyle{ieeetr}
}
\end{document}